\def\year{2022}\relax
\documentclass[letterpaper]{article} 
\usepackage{aaai22}  
\usepackage{times}  
\usepackage{helvet}  
\usepackage{courier}  
\usepackage[hyphens]{url}  
\usepackage{graphicx} 
\usepackage{natbib}  
\usepackage{caption} 
\DeclareCaptionStyle{ruled}{labe`lfont=normalfont,labelsep=colon,strut=off} 
\usepackage{algorithm}
\usepackage{algorithmic}
\usepackage{subcaption}
\usepackage{hyperref}

\usepackage{amssymb}
\usepackage{pifont}
\newcommand{\cmark}{\ding{51}}%
\newcommand{\xmark}{\ding{55}}%

\usepackage{newfloat}
\usepackage{listings}
\floatstyle{ruled}
\newfloat{listing}{tb}{lst}{}
\floatname{listing}{Listing}
\usepackage{multirow}
\usepackage{xcolor}
\usepackage{amsmath}
\usepackage[frozencache=true,cachedir=mintedcache]{minted}

\nocopyright

\title{Higher Order Recurrent Space-Time Transformer for Video Action Prediction}

\author{
    Tsung-Ming Tai\textsuperscript{\rm 1},
    Giuseppe Fiameni\textsuperscript{\rm 1},
    Cheng-Kuang Lee\textsuperscript{\rm 1},
    Oswald Lanz \textsuperscript{\rm 2}
}
\affiliations{
    \textsuperscript{\rm 1} NVIDIA AI Technology Center\\
    \textsuperscript{\rm 2} Fondazione Bruno Kessler\\
    ntai@nvidia.com, 
    gfiameni@nvidia.com,
    cklee@nvidia.com,
    lanz@fbk.eu
}

\begin{document}

\maketitle

\begin{abstract}
Endowing visual agents with predictive capability is a key step towards video intelligence at scale. The predominant modeling paradigm for this is sequence learning, mostly implemented through LSTMs. Feed-forward Transformer architectures have replaced recurrent model designs in ML applications of language processing and also partly in computer vision. In this paper we investigate on the competitiveness of Transformer-style architectures for video predictive tasks. To do so we propose HORST, a novel higher order recurrent layer design whose core element is a spatial-temporal decomposition of self-attention for video. HORST achieves state of the art competitive performance on Something-Something early action recognition and EPIC-Kitchens action anticipation, showing evidence of predictive capability that we attribute to our recurrent higher order design of self-attention.
\end{abstract}

\section{Introduction}
\label{sec:intro}
Recognizing human actions from videos is a 
widely studied problem in computer vision. Most work has addressed action recognition as a video clip classification problem, lately almost exclusively with deep learning approaches, and a steady progress has 
pushed the limits of spatio-temporal feature learning from video. An underlying assumption in action recognition as clip classification is that of complete and synchronous observation, that is, the action to be recognized is immediately accessible and entirely represented in the 
clip. Such completeness assumption no longer holds with 
action prediction tasks, which is to forecast the future from 
observations of the past. In 
practice observations may be streamed and 
elaborated progressively to perform and revise future prediction continuously over time. These are requirements shared among many real world applications of video based 
prediction, e.g. in human-robot collaboration, real-time video surveillance, and autonomous driving.

\begin{figure}\centering
  \includegraphics[width=\columnwidth]{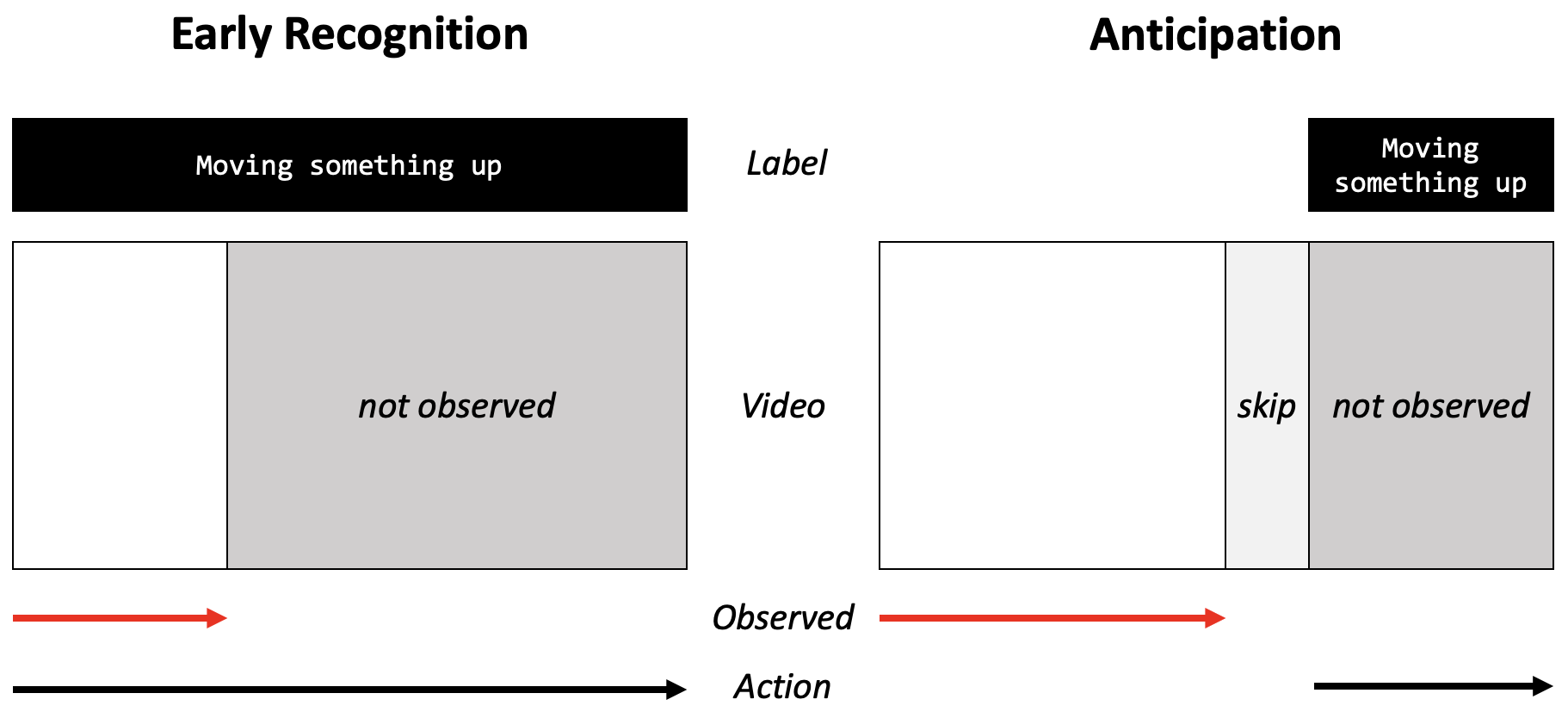}
  \caption{Early Recognition vs Anticipation. In early recognition the action can be partially observed, in anticipation it can only be inferred as a consequence of the observed.}
  \label{fig:early-vs-anticipation}
\end{figure}

We illustrate the problem setting of early action recognition and action anticipation, which are the two major video action prediction tasks, in Figure~\ref{fig:early-vs-anticipation}. As opposed to recognition, both require prediction of action labels beyond the extent of the observed video sub-clip. In early recognition, the target label is global and the action is already represented, even if not to its full extent, as a signal in the sub-clip. In anticipation, the target action only stays in causal relation to the signal in the sub-clip, but is not directly observable in it. It must be forecast as one possible consequence of the observed. While a clip classification design could be deployed for prediction, such an approach would not take advantage of these peculiar challenges that are not necessarily relevant for recognition. The problem to solve is different.
 
A standard modeling framework for anticipation and early recognition is recursive sequence prediction~\cite{DBLP:conf/nips/SuBKHKA20,furnari2020rulstm,qiwang2021srl}, where video frames are consumed in sequence to progressively update a representation of the relevant video content. 
Recent work has expanded upon LSTM or GRU design to realize models capable of capturing higher order correlations across time~\cite{DBLP:conf/nips/SuBKHKA20}, that fork parallel models at each iteration to simultaneously update a representation of the past while predicting the future with an encoder-decoder approach~\cite{furnari2020rulstm}, and revise intermediate representations through a dynamic re-weighting mechanism in a self-regulated learning framework~\cite{qiwang2021srl}. The shared objective of these works is to mitigate performance degradation due to accumulation of anticipation error. Some works furthermore attempt to tame the  uncertainty induced by longer anticipation time through predicting multiple futures~\cite{fartha2020cvprw} and by enforcing bidirectional predictions through cycle consistency~\cite{DBLP:journals/corr/abs-2009-01142}.

In this paper we explore the design and effective learning of space-time transformers for predictive tasks such as early recognition and anticipation. Transformers have replaced recurrent models in ML applications of language processing and also partly in computer vision. Latest work demonstrates that convolution-free, pure transformer architectures can compete with video CNNs on action classification~\cite{DBLP:journals/corr/abs-2102-05095,DBLP:journals/corr/abs-2103-15691}. Timely enough, in this paper we investigate whether they can replace LSTMs on video predictive tasks, where LSTMs are still the common modeling paradigm of best performing methods. To verify this, we propose a novel higher order recurrent layer whose core element is a spatial-temporal decomposition of self-attention for video. It is higher order as it maintains a state queue in the attention mechanism to keep track of previously recorded information. It is recurrent in the way the queue is updated at each time step. Our layer is lightweight and has a transparent design, and achieves state of the art competitive performance 
when deployed for early action recognition and anticipation. 


\section{Related Work}
\label{sec:related}
 
\subsubsection{Action Recognition.}
Classic approaches
extract hand-crafted features from video and train a classifier on these to solve the task~\cite{DBLP:journals/ijcv/Laptev05,DBLP:journals/ijcv/WangKSL13}. Modern approaches blend feature extraction and classification into modular, end-to-end trainable neural architectures.
Building on the success of 2D CNNs for image recognition, early approaches use temporal pooling of frame-level features to process video as a set of images~\cite{DBLP:conf/eccv/WangXW0LTG16,DBLP:conf/cvpr/GirdharRGSR17} or use two-stream architectures to fuse frame features with features extracted from 
flow~\cite{DBLP:conf/nips/SimonyanZ14,DBLP:conf/cvpr/FeichtenhoferPZ16}. 3D CNNs process videos in space-time by expanding 2D kernels along the temporal dimension~\cite{DBLP:conf/iccv/TranBFTP15,DBLP:conf/cvpr/CarreiraZ17,DBLP:conf/cvpr/Feichtenhofer20}, requiring more parameters
and compute. Learning spatio-temporal features with less is a shared objective of many recent works. These include decomposing 3D convolutions into 2D spatial followed by 1D temporal \cite{DBLP:conf/cvpr/TranWTRLP18}, replacing 1D with 
time shifts \cite{DBLP:conf/iccv/LinGH19} or 
gating \cite{DBLP:conf/cvpr/SudhakaranEL20}, 
split
channel dimensions using group convolutions \cite{DBLP:conf/iccv/TranWFT19}, modeling interactions between separated dimensions \cite{DBLP:conf/cvpr/LiJSZKW20}, and through adaptive fusion~\cite{DBLP:journals/corr/abs-2102-05775}. Sequence learning models based on LSTM have been augmented with attention~\cite{DBLP:journals/cviu/LiGGJS18,DBLP:conf/nips/GirdharR17,DBLP:conf/cvpr/SudhakaranEL19}.
Transformer inspired attention layers
have been introduced in video CNNs~\cite{DBLP:conf/cvpr/GirdharCDZ19,DBLP:conf/cvpr/0004GGH18,DBLP:conf/nips/ChenKLYF18}, and recent studies show
that pure self-attention based, convolution-free architectures can compete in action recognition~\cite{DBLP:journals/corr/abs-2102-05095, DBLP:journals/corr/abs-2103-15691}.  
 
\subsubsection{Early Action Recognition.}
Works 
often expand upon action recognition architectures. Using a 
classifier trained on sub-clips is utilized as a baseline, which is valid in early recognition since the target action is already observed in the sub-clip. Given the nature of the problem that is to predict from an initial part of a frame sequence, a most intuitive framework can be that of recurrent models such as LSTMs. LSTM architectures can be trained under the assumption that recognition confidence should be non-decreasing as the model observes more of the action by introducing a ranking loss~\cite{DBLP:conf/cvpr/MaSS16}. Following a same assumption, a multi-stage LSTM architecture integrating context-aware and action-aware is trained with a loss that encourages to make correct predictions as early as possible in
the input sequence~\cite{DBLP:conf/iccv/AkbarianSSFPA17}. Sequential context can be leveraged to reconstruct missing information in the features extracted from partial video by learning from fully observed action videos~\cite{DBLP:conf/cvpr/KongTF17}.
LSTMs are the building blocks of a two-stream feedback network where one stream processes the input and the other models the temporal relations to promote the representation of temporal dependencies~\cite{DBLP:conf/wacv/GeestT18}. Eidetic LSTM \cite{DBLP:conf/iclr/WangJYLLF19} introduces an attentive mechanism to memorize local appearance or short-term motion, and use it to improve predictive spatio-temporal feature learning. A higher order LSTM with convolutional gates uses multiple inputs from the past to update its internal state~\cite{DBLP:conf/nips/SuBKHKA20}. Using tensor train decomposition, their extension effectively dominates compute and  parameter count to grow at most linearly in time and space.
 
\subsubsection{Action Anticipation.}
Differently from early recognition, anticipation is to predict the action from observations before it actually starts. The problem has been studied in the context of third person vision but recently also with major efforts in robotic and egocentric, first person vision.
Early work combines a Markov decision process model with ideas from control theory to forecast human trajectories using semantic scene context~\cite{DBLP:conf/eccv/KitaniZBH12}. \cite{DBLP:journals/pami/KoppulaS16} propose a CRF 
framework which integrates object affordances to inform about possible future actions.
\cite{DBLP:conf/cvpr/VondrickPT16} address action anticipation by training a CNN to regress 
representations of future frames from past ones in an unsupervised way. A encoder-decoder LSTM network is presented in~\cite{DBLP:conf/bmvc/GaoYN17a} which takes multiple past representations as
input and learns to anticipate a time series of future representations. \cite{DBLP:conf/cvpr/FarhaRG18} investigate the use of CNN and GRU to learn future video labels based on previously seen content. An anticipatory model is combined with an auxiliary model to 
reason about how actions and scene attributes may co-evolve over time~\cite{DBLP:conf/cvpr/MiechLSWTT19}. 
RU-LSTM~\cite{DBLP:conf/iccv/FurnariF19} is a learning architecture which processes RGB frame snippets, optical flow and object features using two LSTMs and modality attention to anticipate future actions. The two LSTMs behave like a encoder-decoder, where the first progressively summarizes the observed together with the second that unrolls over future predictions without observing. A self-regulated learning framework 
is presented in~\cite{9356220}, that utilizes LSTMs to progressively re-weight previously observed content, and to rectify predicted intermediate representations. 
\cite{fernando2021anticipating} distill the information of the future to the past representation by maximizing the correlation between past and future frames in the feature space for video action prediction.
 
\subsubsection{Higher Order Recurrent Networks.}
Following the first order Markov chain assumption, RNNs can efficiently process sequential inputs by learning a transition model of underlying dynamics. In practice, it is difficult to train RNNs to capture long-term dependency due to vanishing or exploding gradients \cite{bengio1994learning}.  To better capture long term dependencies, \cite{soltani2016higher} keep track of more past states in a higher order recurrent network. Further studies analyze different aggregation functions to recur on the set of past states~\cite{yu2017long}.
\cite{DBLP:conf/nips/SuBKHKA20} derive a tensor-train decomposition of higher order convolutional LSTMs for video predictive tasks.
 
\subsubsection{Vision Transformers.}
Vision Transformer (ViT) \cite{dosovitskiy2020image} can handle the input spatial dimension by splitting the inputs into 16x16 patches. However, taking full advantage of the additional temporal dimension in video is still challenging. \cite{DBLP:journals/corr/abs-2102-05095} explores different self-attention schemes to incorporate spatial and temporal information for video recognition. \cite{DBLP:journals/corr/abs-2103-15691} further factorizes the spatial-temporal structure by deploying separated multi-head attentions for each spatial and temporal dimension. \cite{girdhar2021anticipative} propose a vision transformer based architecture for future action prediction with causal attention design. A distinguishing element in our work is that we do not apply patching and flattening to the input; we process the input at its full spatial resolution without imposing any structure prior (grid). Our custom decomposition of self-attention is also tightly integrated with a recurrent design to update the representation for future prediction from a queue of past representations.  


\section{Preliminaries}
 
\subsubsection{Higher Order Recurrence.}
Conventional recurrent networks, where the output of step $t$ only depends on its previous $t-1$ state, follow a \textit{first order} Markov assumption
\begin{equation}
    h_t = g(x_t, h_{t-1})
\label{eq:fornn}
\end{equation}
where $g$ can be any arbitrary nonlinear function including LSTM and GRU, $x_t$ is the current input and $h_t,h_{t-1}$ are current and previous hidden state. Extending to \textit{higher order} Markov assumption, where the output of step $t$ depends on multiple historical states $t-1$ to $t-S$ with $S$ indicating the \textit{order}, equation \eqref{eq:fornn} can be reformulated as
\begin{eqnarray}
    h_t = g(x_t, \phi(h_{t-1:t-S}))
\label{eq:hornn}
\end{eqnarray}
where $\phi$ is an aggregation function for a state queue $h_{t-1:t-S}$. Various aggregation functions have been proposed, such as linear \cite{soltani2016higher}, polynomial \cite{yu2017long}, and tensor-train \cite{DBLP:conf/nips/SuBKHKA20}.

\begin{figure}
\centering
  \includegraphics[width=.9\columnwidth]{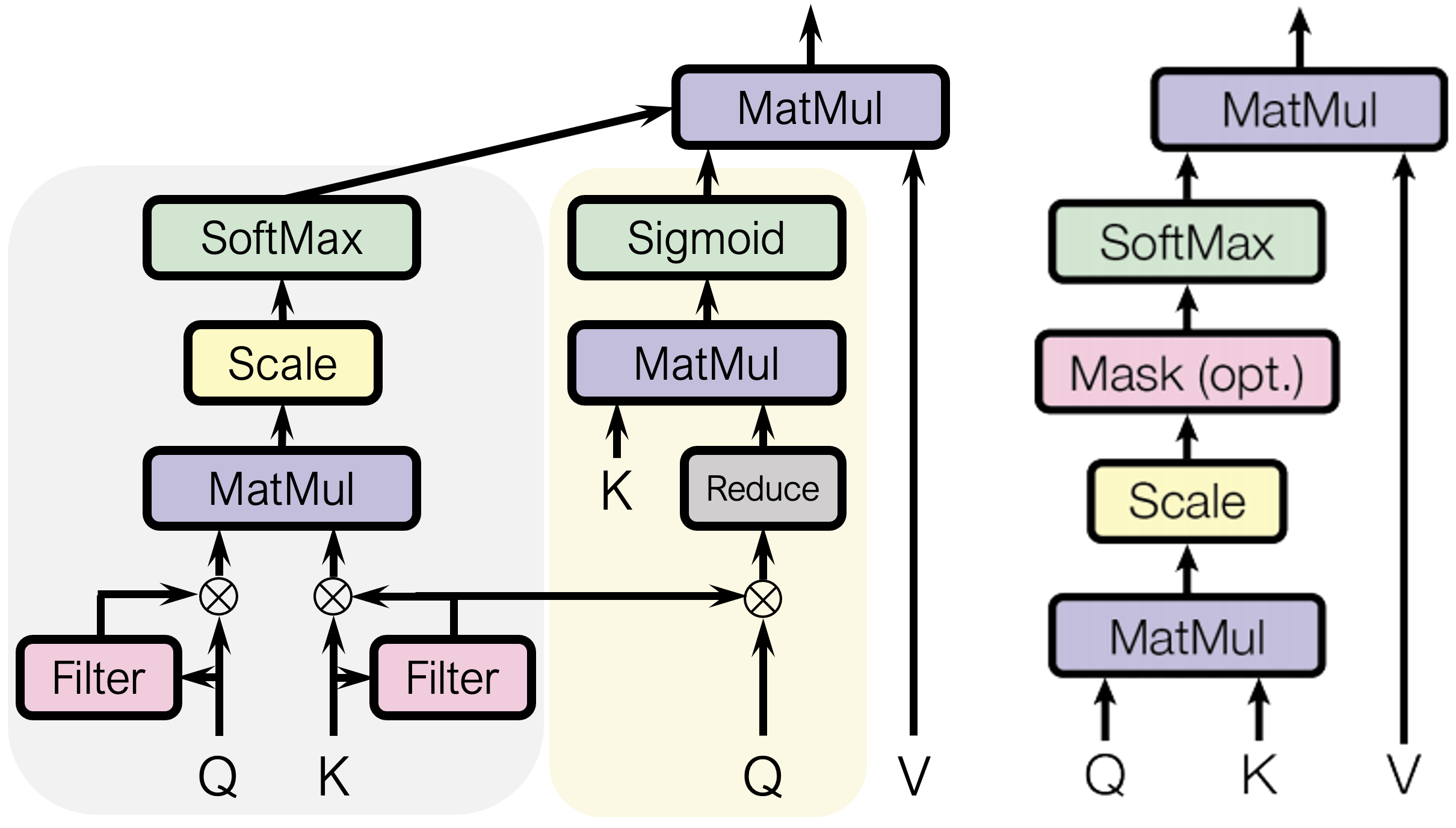}
  \caption{Left: Proposed Spatial-Temporal Attention. (Temporal branch is in grey area, and spatial branch is in yellow); Right: Attention in \cite{DBLP:conf/nips/VaswaniSPUJGKP17}.}
  \label{fig:statt_vs_sa}
\end{figure}

\begin{figure*}
\centering
  \includegraphics[width=\textwidth]{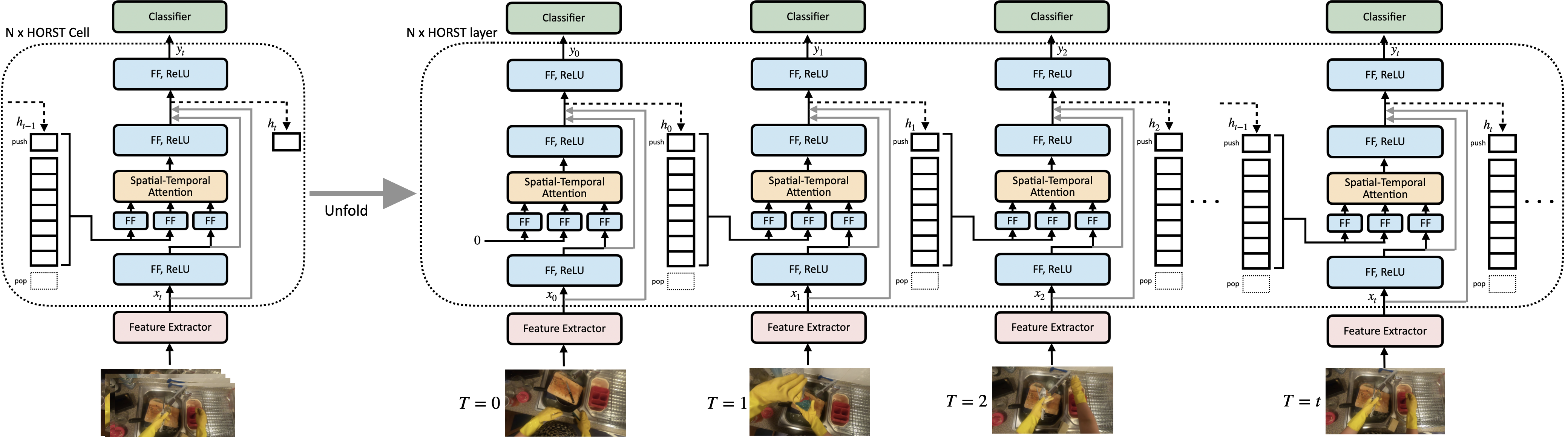}
  \caption{An overview of HORST for video prediction, in recurrent and unfold view.}
  \label{fig:horst_unfold}
\end{figure*}

 
 
\subsubsection{Attention.}
The attention proposed in \cite{DBLP:conf/nips/VaswaniSPUJGKP17} takes as input query $Q$, key $K$, and value $V$. The outcome is obtained by evaluating the dot-product between $Q$ and $K$, scaled by a softmax function, and multiplying it to $V$
 
\begin{equation}
    \text{ATT}(Q, K, V) = \text{softmax}(\frac{Q^TK}{\sqrt{C}})V
\end{equation}
where $C$ is a scaling factor. The attention was designed for vector inputs. It is parameter-free, and the learning capacity in transformers is with the linear embeddings that map a given layer input $x$ to the three attention inputs $Q,K,V$.

\section{Method}
\subsection{Spatial-Temporal Attention}
To exploit the effective information in spatial-temporal structure, we propose a light-weighted and computation-efficient attention, which integrates spatial and temporal operators from separated branches. Figure~\ref{fig:statt_vs_sa} compares the proposed spatial-temporal attention with the attention in \cite{DBLP:conf/nips/VaswaniSPUJGKP17}. To define spatial and temporal branch operators, we introduce \textit{spatial filter} maps $f_\mathcal{K}, f_\mathcal{Q}$ for keys and query. Their role is attending the relevant spatial regions in keys and query. We use the following
\begin{equation}
    f_{\mathcal{X}}(X) = \text{sigmoid}(\theta_{\mathcal{X}} * [X_{max}, X_{avg}] + b_{\mathcal{X}})
    \label{eq:filter}
\end{equation}
where $*$ is convolution, $X_{avg}, X_{max}$ are channel mean and max pooled, $\theta_\mathcal{K}, \theta_\mathcal{Q}$ and $b_{\mathcal{K}}, b_{\mathcal{Q}}$ are convolution kernels and biases, and sigmoid is to map to range [0:1]. We use $f_\mathcal{K},f_\mathcal{Q}$ as weight maps to filter keys and query for space-time attention.

 
{\em Spatial branch:} $\mathcal{S}(Q, K)$ provides pixel-wise weight maps for spatial attention at each timestep. Weight maps are measured by the sigmoid of dot-product between keys and a global average pooling of filtered key-query comparisons \begin{eqnarray}
\hat{Q} &=& \text{GlobalAveragePool}(f_{\mathcal{K}}(K) \cdot Q)\nonumber\\
\mathcal{S}(Q, K) &=& \text{sigmoid}(\hat{Q}^T K) \label{eq:spatial}
\end{eqnarray}
where $\cdot$ is element-wise multiplication.

{\em Temporal branch:} $\mathcal{T}(Q, K)$ measures the importance weights in the temporal dimension. The weights are calculated by dot-product attention between self-filtered keys and query, followed by scaled softmax
\begin{eqnarray}
\mathcal{T}(Q, K) = \text{softmax}(\frac{(f_{\mathcal{Q}}(Q)\cdot Q)^T (f_{\mathcal{K}}(K) \cdot K)}{\sqrt{C}})
\label{eq:temporal}
\end{eqnarray}
 
{\em Spatial-temporal attention:} Combining spatial and temporal branches, defined in \eqref{eq:spatial} and \eqref{eq:temporal}, the proposed spatial-temporal decomposition of self-attention is
\begin{eqnarray}
\text{STATT}(Q, K, V) = (\mathcal{S}(Q, K) \otimes \mathcal{T}(Q, K)) \cdot V
\label{eq:statt}
\end{eqnarray}
where $\otimes$ is cross-product. In this model, keys and query are utilized differently in spatial and temporal attention computation as they get pre-filtered by the learnable $f_\mathcal{K},f_\mathcal{Q}$. Hence, our formulation of self-attention is not parameter-free.

{\em Ops counts:} Our spatial-temporal design requires $(6S+1)CHW + 18(S+1)HW$ ops, where $S$ is the key-value sequence length. This is far less than $2S(HW)^2C$ of joint space-time attention \cite{DBLP:journals/corr/abs-2102-05095}, and approximately twice the ops of full-temporal, $2SHWC$, while providing a more fine-grained attention design.
 
\subsection{Higher Order Recurrent Space Time Transformer}
 
Figure~\ref{fig:horst_unfold} shows an illustration of HORST architecture and its unfolding to process video inputs.
All 
$\texttt{FF}$ blocks
are \texttt{Conv-LayerNorm} transformations.
At each step $t$, we feed the video frame through a 2D-CNN backbone to obtain feature map $x_t$, and encode it with $\texttt{FF}_x$ to obtain the intermediate representation $e_t$. We then project $e_t$ to $Q,K,V$ independently using $\texttt{FF}_{Q},\texttt{FF}_{K},\texttt{FF}_{V}$. The spatial-temporal attention output from \eqref{eq:statt} is then projected to $h_t$ using $\texttt{FF}_h$. Layer output $y_t$ is computed from $h_t$ through a final $\texttt{FF}_y$ informed by shortcuts $x_t, e_t$. The new state $h_t$ is pushed to the queue and the oldest state $h_{t-S}$ is released.

By instantiating $\phi$ in \eqref{eq:hornn} with spatial-temporal attention from \eqref{eq:statt}, a HORST layer can be formalized as
\begin{eqnarray}
e_t \!\!\!&=&\!\!\! \texttt{ReLU}(\texttt{FF}_x(x_t)\label{eq:encoded_x})\\
h_t \!\!\!&=&\!\!\! \texttt{ReLU}(\texttt{FF}_{h}(e_t + \text{STATT}(Q_t, K_{t-1:t-S}, V_{t-1:t-S})))\nonumber\\
y_t \!\!\!&=&\!\!\! \texttt{ReLU}(\texttt{FF}_{y}(h_t + x_t))\nonumber
\end{eqnarray}
with
\vspace{-.05in}
\begin{eqnarray}
Q_t &=& \texttt{FF}_Q(e_t)\label{eq:q}\\
K_{t-1:t-S} &=& \texttt{FF}_K([Q_{t-1:t-S}; h_{t-1:t-S}])\label{eq:k}\\
V_{t-1:t-S} &=& \texttt{FF}_V([Q_{t-1:t-S}; h_{t-1:t-S}])\label{eq:v}
\end{eqnarray}
where $[.;.]$ is concatenation.
 
HORST can be viewed as a transformer that processes sequence data in a recurrent manner with an internal memory queue managed by a first-in first-out policy. Having spatial-temporal attention as aggregation function $\phi$ in \eqref{eq:hornn} makes HORST capable of accessing specific memory states while skipping others. Furthermore, our attention design is single-head. Multi-head tends to produce over-smoothed temporal weighting. We found empirically that the performance of HORST is stronger with single-head, considering the number of states is small.
In our ablation study, we also show that position encoding, which is widely adopted in transformer, is not needed in HORST given explicit recurrence. By attaching $e_t$ in \eqref{eq:encoded_x} to keys and values, the temporal position information is preserved.

\section{Experiments}
\label{sec:experiment}
\begin{figure*}[!ht]
\centering
    \begin{minipage}{\textwidth}
    \centering
    \includegraphics[width=0.75\textwidth]{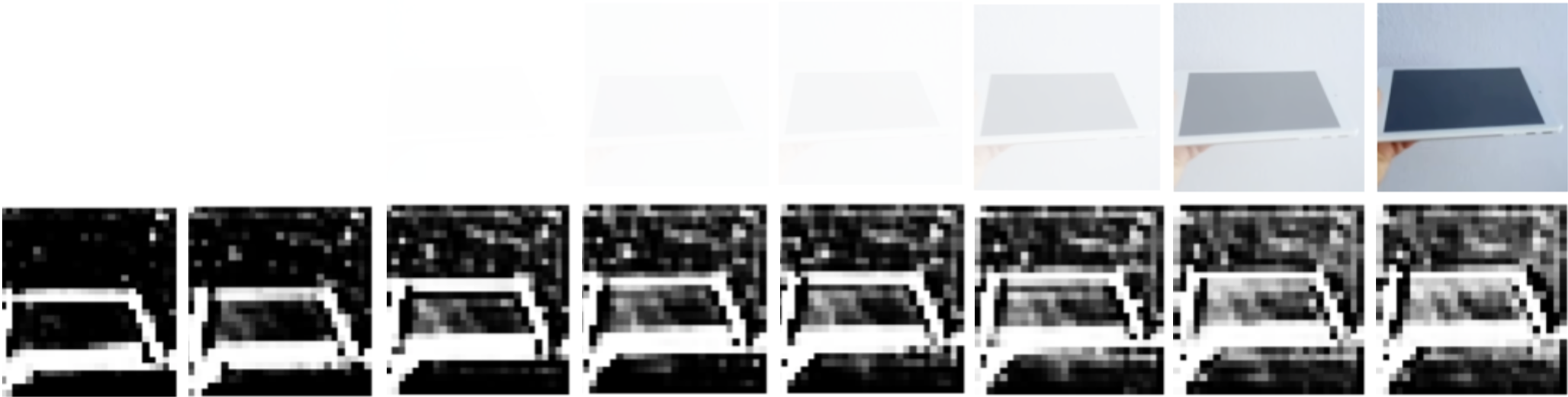}
    \subcaption[]{\textbf{Ground Truth}: Moving [something] up. \textbf{Prediction}: Moving [something] up. (\textcolor{green}{\cmark})}
    \label{fig:ssv2_visa}
    \end{minipage}\vspace{.03in}
    
    \begin{minipage}{\textwidth}
    \centering
    \includegraphics[width=0.75\textwidth]{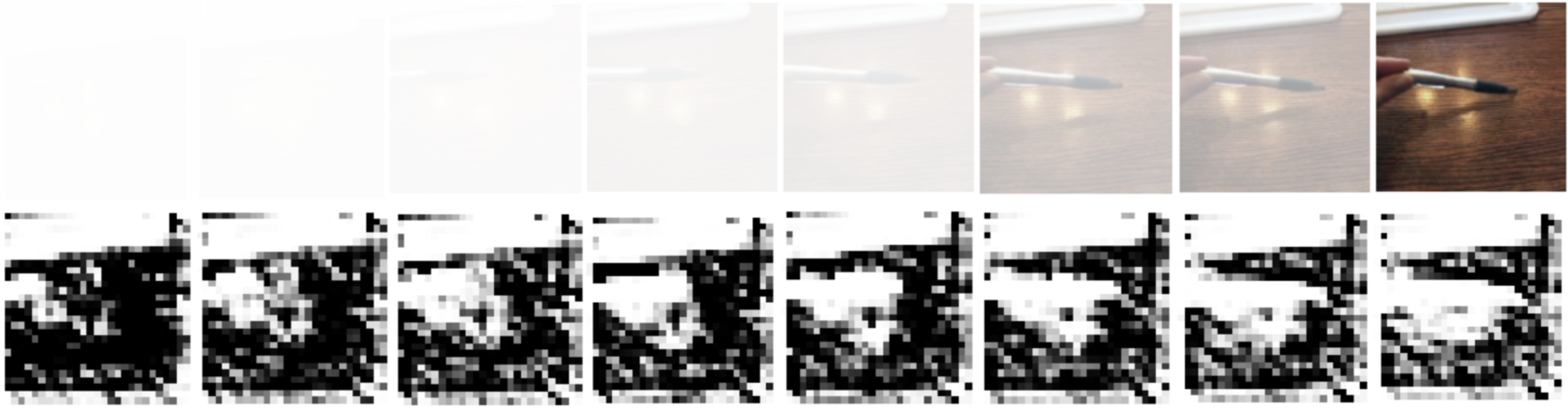}
    \subcaption[]{\textbf{Ground Truth}: Putting [something] on a flat surface without letting it roll. \textbf{Prediction}: Putting [something] on a surface. (\textcolor{red}{\xmark})}
    \label{fig:ssv2_visb}
    \end{minipage}
    
\caption{Visualization samples on SSv2: we show input frames weighted by temporal attention and spatial attention maps from the final $S=8$ timesteps leading to the prediction. Captions reveal ground-truth annotation and top-1 prediction.}
\label{fig:ssv2_vis}
\end{figure*}
 
\subsubsection{Something-Something v2 (SSv2).}
SSv2 \cite{DBLP:conf/iccv/GoyalKMMWKHFYMH17} is a large collection of action video clips, with duration ranging from 2 to 6 seconds. We follow the data split scheme in \cite{DBLP:conf/iclr/WangJYLLF19} for early recognition, with a subset of 41 categories. There are a total 56769 video clips for training and 7503 for validation.
 
\subsubsection{EPIC-Kitchens (EK55, EK100).} 
EK55 \cite{atsn} is a large scale egocentric video dataset, captured by 32 subjects in 32 kitchens. The action anticipation split is inherited from \cite{DBLP:conf/iccv/FurnariF19} and amounts to 23492 action segments for training and 4979 for validation. Videos are categorized into 125 verbs and 352 nouns. All unique verb-noun pairs define 2513 action categories. EK100 \cite{Damen2020RESCALING} extends EK55 from 55 to 100 hours of video. EK100 considers 97 verbs and 300 nouns. Unique verb-noun pairs define 3087 action categories.

\subsection{Implementation Details}
\subsubsection{Architecture.} 
In early recognition on SSv2 dataset, two 3x3 convolution layers with 128 filters and stride 2 are stacked as backbone extractor, which shrinks the inputs from 224x224 to 56x56. Four HORST layers, with strides [1, 1, 2, 2], take the feature maps from backbone and produce 14x14 outputs. A classifier 
then transforms the HORST outputs into the target action predictions. In anticipation with EK55 and EK100 datasets, we use the pretrained BN-Inception model of \cite{DBLP:conf/iccv/FurnariF19} as backbone. All input frames are resized to 256x454. The feature maps before the global average pooling in BN-Inception are extracted and fed through two HORST layers followed by a classifier 
forming the noun, verb, and action predictions. We also experiment with a task specific classifier design, where we adopt the unrolling LSTM from \cite{DBLP:conf/iccv/FurnariF19}. The unrolling classifier unfolds the final HORST output over the unobserved interval with an LSTM, till the moment where the action is expected to start, and then computes the predictions.

\subsubsection{Training.} 
We adopt RandAugment~\cite{cubuk2020randaugment} in all experiments. We use AdaBelief \cite{DBLP:conf/nips/ZhuangTDTDPD20} in combination with look-ahead optimizer \cite{NEURIPS2019lookahead}. Weight decay is set to 0.001. Learning rate is initially set to 0.002 and cosine annealed to 0 on the final 25\% of epochs. The total training epochs is set to 50. We use 4 $\times$ NVIDIA V100 32GB GPUs for training. Batch size is set to 8 for early recognition and 32 for anticipation.

\begin{figure*}[!ht]
\centering
    \begin{minipage}{\textwidth}
    \centering
    \includegraphics[width=0.8\textwidth]{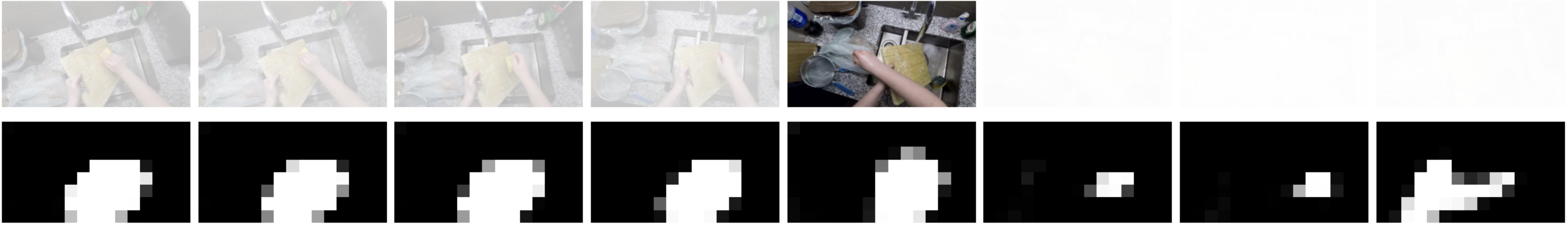}
    \subcaption[]{\textbf{Ground Truth}: \textcolor{green}{Wash Board}; \textbf{Top-5 Predictions}: \textcolor{green}{Wash Board}, \textcolor{red}{Put-Down} \textcolor{green}{Board}, \textcolor{red}{Turn-Off Tap}, \textcolor{red}{Put-Down Sponge}, \textcolor{red}{Stir Pan}.}
    \label{fig:ek55_visa}
    \end{minipage}\vspace{.03in}
    
    \begin{minipage}{\textwidth}
    \centering
    \includegraphics[width=0.8\textwidth]{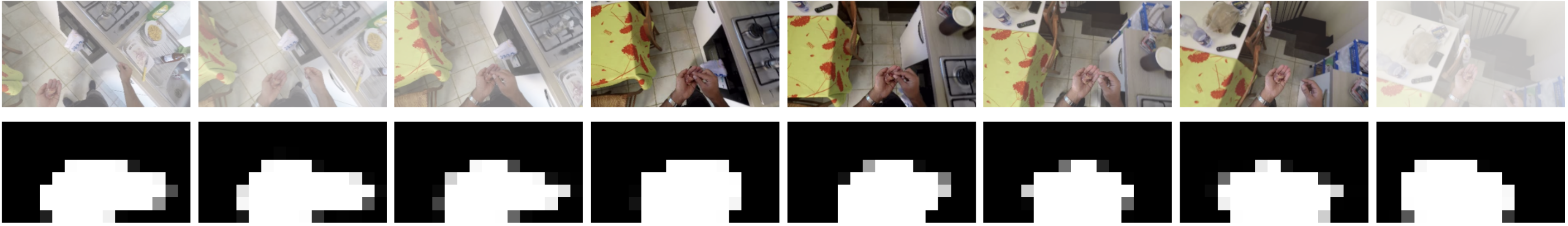}
    \subcaption[]{\textbf{Ground Truth}: \textcolor{green}{Throw Onion}; \textbf{Top-5 Predictions}: \textcolor{red}{Open Door}, \textcolor{green}{Throw} \textcolor{red}{Skin}, \textcolor{red}{Take} \textcolor{green}{Onion}, \textcolor{red}{Take Skin}, \textcolor{red}{Take-Off Skin}.}
    \label{fig:ek55_visb}
    \end{minipage}
    
\caption{Visualization samples on EK55: we show input frames weighted by temporal attention and spatial attention maps from the final $S=8$ timesteps leading to the predictions. Captions reveal ground-truth annotation and top-5 predictions. 
}
 
\label{fig:ek55_vis}
\end{figure*}

\subsection{Early Action Recognition}
\subsubsection{Sampling.} We conduct early recognition on SSv2 dataset, when only the initial 25\% or 50\% of total frames are observed. Inputs are resized to 224x224 and all frames (24 fps) are used for prediction. Each sample is variable length.

\subsubsection{Quantitative Results.}
Table~\ref{tab:ssthv2_25_and_50er_acc} shows the top-1 accuracy compared with state-of-the-art methods in both 25\% and 50\% early recognition settings. All the recurrent-based methods outperform the 3D-CNN baseline. Conv-TT-LSTM is a 3-order ConvLSTM model with tensor-train decomposition. We reproduced Conv-TT-LSTM in our training protocol (e.g., augmentations, optimizer, and learning rate scheduling) resulting in significant improvements. HORST further achieves state-of-the-art performance by a large margin on both 25\% and 50\% settings, with +4.9\% and +5.5\% higher top-1 action accuracy than Conv-TT-LSTM.
 \begin{table}[!ht]
\caption{SSv2 25\% and 50\% early action recognition results. $\dagger$ indicates reproduction by us using our training protocol.}
    \centering
    \small
    \begin{tabular}{l|c|c}
         \multirow{2}{*}{Methods} &
         \multicolumn{2}{c}{Top-1 Accuracy(\%)} \\
         & 25\% frames & 50\% frames \\
         \hline
         3D-CNN & 13.3 & 20.7 \\
         E3D-LSTM \cite{DBLP:conf/iclr/WangJYLLF19} & 14.6 & 22.7 \\
         ConvLSTM & 15.5 & 22.0 \\
         Conv-TT-LSTM \cite{DBLP:conf/nips/SuBKHKA20} & 19.5 & 30.1 \\ 
         \hline
         Conv-TT-LSTM$\dagger$ & 29.3 & 42.3 \\
         HORST (Ours) & \textbf{33.9} & \textbf{47.8} \\
         
    \end{tabular}
    \label{tab:ssthv2_25_and_50er_acc}
\end{table}

\subsubsection{Visualization Analysis.}
Figure~\ref{fig:ssv2_vis} visualizes temporal (transparency on input frames) and spatial attention (second row) for two samples from the SSv2 validation set.
Figure~\ref{fig:ssv2_visa} is a sample with the ground truth \texttt{Moving [something] up}. Spatial attention captures the border of the object whose vertical translation is clearly associated to the target action. Temporal attention focuses on the few recent states due to the steady movement of the sample. Figure~\ref{fig:ssv2_visb} is an example with incorrect prediction. Without observing the subsequent frames, the model can hardly decide whether the object will be rolling or not. It is worth noting that in this sample, spatial-temporal attention produces the highest temporal weights when the object touched the surface, which is an important event for reasonable prediction.

\subsection{Ablation Study}
 
Ablations are all performed on SSv2 25\% early recognition.
 
\subsubsection{Spatial-Temporal Attention.}
We compare different attention variants in Table~\ref{tab:ssthv2_ablation_layer}.  Temporal only attention 
is with the constant
$\mathcal{S}(Q,K) = 1$ instead of \eqref{eq:spatial}.
Spatial only attention 
is with the constant
$\mathcal{T}(Q,K) = \frac{1}{S}$ instead of \eqref{eq:temporal}.
Spatial-temporal is our proposed attention using \eqref{eq:temporal} and \eqref{eq:spatial}.
\begin{table}[!ht]
\caption{Ablation on attention design}
    \centering
    \small
    \begin{tabular}{l|c}
         Attention & Top-1 Accuracy (\%) \\
         \hline
         $4 \times \text{Temporal only}$ & 32.7 \\
         $4 \times \text{Spatial only}$ & 31.2 \\
         $4 \times \text{Spatial-Temporal}$ & \textbf{33.9} \\
         \hline
         $2 \times \text{Joint Space-Time}$ & 30.4 \\
         $2 \times \text{Spatial-Temporal}$ & \textbf{32.2} \\
    \end{tabular}
    \label{tab:ssthv2_ablation_layer}
\end{table}

\begin{table}[!ht]
\caption{Impact of model order}
    \centering
    \small
    \begin{tabular}{l|c}
         Order & Top-1 Accuracy (\%) \\
         \hline
         $S=1$ & 30.7 \\
         $S=3$ & 31.2 \\
         $S=8$ & \textbf{33.9} \\ 
         $S=12$ & 33.8 \\ 
         $S=16$ & 32.7 \\ 
    \end{tabular}
    \label{tab:ssthv2_ablation_orders}
\end{table}
 
\begin{table}[!ht]
\caption{Ablation on choices for key and value 
}
    \centering
    \scriptsize
    \begin{tabular}{l|l|c}
         Key ($K$) & Value ($V$) & Acc (\%) \\
         \hline
         $\texttt{FF}_K(e_{t-1:t-S})$ & $\texttt{FF}_V(h_{t-1:t-S})$ & 30.6 \\
         $\texttt{FF}_K([e_{t-1:t-S}; h_{t-1:t-S}])$ & $\texttt{FF}_V([e_{t-1:t-S}; h_{t-1:t-S}])$ & \textbf{33.9} \\
    \end{tabular}
    \label{tab:ssthv2_ablation_queue_concat}
\end{table}
The performance differences with temporal only and spatial only imply that both local spatial and temporal weighting using filtered state information is beneficial and best exploited when combined.
We compare our spatial-temporal attention with joint space-time attention from \cite{DBLP:journals/corr/abs-2102-05095} using two layers\footnote{Because of heavy memory consumption, we cannot use four joint space-time attention layers with the same batch size. Lowering the batch size degrades its accuracy.}. Ours achieves an absolute +1.8\% gain over joint space-time with same number of layers. Using four HORST layers instead, gain raises to +3.5\%.


\subsubsection{Order.}
Table~\ref{tab:ssthv2_ablation_orders} compares model performance under different orders $S$. We observe the best accuracy when $S=8,12$. The drop in performance with $S = 16$ may relate to the gradient instability issue reported in \cite{DBLP:conf/nips/SuBKHKA20}\footnote{HORST can still enjoy higher order design with peak performance at $S=8$, instead of $S=3$ reported in prior works.}. Our higher order design shows significant improvement over the first order variant, achieving up to +3.2\% higher accuracy.

\subsubsection{Position Encoding.}
We compare two different choices for spatial-temporal attention inputs. Considering that spatial-temporal aggregates past states with different associated weights for each time step, first choice in the Table~\ref{tab:ssthv2_ablation_queue_concat} causes key-value misaligned in time. Binding both key and value with $e_t$ (which is directly encoded from the layer input $x_t$) provides the temporal position encoding to each time step in both $K$ and $V$. Indeed, this leads to a +3.3\% improvement.
 
\begin{table*}[!ht]
\caption{EK55 Action Anticipation validation results using RGB. All methods on the table are based on the same BN-Inception with TSN backbone, from \cite{furnari2020rulstm}. HORST-url applies unrolling on top of HORST.}
    \centering
    \small
    \scalebox{0.9}{
    \begin{tabular}{l|cccccccc|ccc|ccc}
    \multirow{2}{*}{Methods} &
    \multicolumn{8}{c}{Top-5 Accuracy (\%) at different $\tau_a$} &
    \multicolumn{3}{c}{Top-5 Acc. (\%) @ 1s} &
    \multicolumn{3}{c}{Mean Top-5 Recall (\%) @ 1s} \\
    & 2 & 1.75 & 1.5 & 1.25 & 1.0 & 0.75 & 0.5 & 0.25 & Verb & Noun & Action & Verb & Noun & Action \\
    \hline
      
    DMR & - & - & - & - & 16.86 & - & - & - & 73.66 & 29.99 & 16.86 & 24.50 & 20.89 & 03.23 \\
    ATSN & - & - & - & - & 16.29 & - & - & - &  77.30 & 39.93 & 16.29 & 33.08 & 32.77 & 07.06 \\
    MCE & - & - & - & - & 26.11 & - & - & - & 73.35 & 38.86 & 26.11 & 34.62 & 32.59 & 06.50 \\
    VN-CE & - & - & - & - & 17.31 & - & - & - & 77.67 & 39.50 & 17.31 & 34.05 & 34.50 & 07.73 \\
    SVM-TOP3 & - & - & - & - & 25.42 & - & - & - & 72.70 & 28.41 & 25.42 & 41.90 & 34.69 & 05.32 \\
    SVM-TOP5 & - & - & - & - & 24.46 & - & - & - & 69.17 & 36.66 & 24.46 & 40.27 & 32.69 & 05.23 \\
    VNMCE+T3 & - & - & - & - & 25.95 & - & - & - & 74.05 & 39.18 & 25.95 & 40.17 & 34.15 & 05.57 \\
    VNMCE+T5 & - & - & - & - & 26.01 & - & - & - & 74.07 & 39.10 & 26.01 & 41.62 & 35.49 & 05.78 \\
    ED & 21.53 & 22.22 & 23.20 & 24.78 & 25.75 & 26.69 & 27.66 & 29.74 & 75.46 & 42.96 & 25.75 & 41.77 & 42.59 & 10.97 \\
    FN & 23.47 & 24.07 & 24.68 & 25.66 & 26.27 & 26.87 & 27.88 & 28.96 & 74.84 & 40.87 & 26.27 & 35.30 & 37.77 & 06.64 \\
    RL & \textbf{25.95} & 26.49 & 27.15 & 28.48 & 29.61 & 30.81 & 31.86 & 32.84 & 76.79 & 44.53 & 29.61 & 40.80 & 40.87 & 10.64 \\
    EL & 24.68 & 25.68 & 26.41 & 27.35 & 28.56 & 30.27 & 31.50 & 33.55 & 75.66 & 43.72 & 28.56 & 38.70 & 40.32 & 08.62 \\
    RU-RGB & 25.44 & 26.89 & \textbf{28.32} & 29.42 & 30.83 & 32.00 & 33.31 & 34.47 & - & - & 30.83 & - & - & - \\
    
    \hline
    HORST & 25.38 & 26.37 & 27.82 & 29.16 & 30.69 & 31.54 & 32.52 & 33.45 & 77.67 & 46.34 & 30.69 & 36.54 & 44.33 & 10.94 \\
    HORST-url & \textbf{25.95} & \textbf{27.03} & 28.24 & \textbf{29.81} & \textbf{31.58} & \textbf{32.68} & \textbf{34.21} & \textbf{35.56} & \textbf{78.80} & \textbf{46.54} & \textbf{31.58} & \textbf{42.62} & \textbf{45.68} & \textbf{12.18} \\
    
    \end{tabular}
    } 
    \label{tab:ek55}
\end{table*}
 
\begin{table}[!ht]
\caption{EK55 Action Anticipation validation results using RGB with top-1 and top-5 action accuracy at $\tau_a = 1\textit{s}$.}
    \centering
    \small
    \begin{tabular}{l|l|l|cc}
         Method & Backbone & Pretrain & Top-1 (\%) & Top-5 (\%) \\
         \hline
         RU-RGB & BNInc & In1k & 13.1 & 30.8 \\
         ActionBanks & BNInc & In1k & 12.3 & 28.5 \\
         ImagineRNN & BNInc & In1k & 13.7 & 31.6 \\
         AVT-h & BNInc & In1k & 13.1 & 28.1 \\
         AVT-h & AVT-b & In21+1k & 12.5 & 30.1 \\
         AVT-h & irCSN152 & IG65M & 14.4 & 31.7 \\
         \hline
         HORST & BNInc & In1k & 12.6 & 30.7 \\
         HORST-url & BNInc & In1k & 12.8 & 31.6 \\
    \end{tabular}
    \label{tab:ek55_top1}
\end{table}
 
\begin{table}[!ht]
\caption{EK100 Action Anticipation validation results using RGB with mean top-5 recall (\%) at 1\textit{s}.}
    \centering
    \small
    \begin{tabular}{l|l|l|ccc}
         Method & Backbone & Pretrain & Verb & Noun & Action \\
         \hline
         RU-RGB & BNInc & In1k & 27.5 & 29.0 & 13.3 \\
         AVT-h & BNInc & In1k & 27.3 & 30.7 & 13.6 \\
         AVT-h & AVT-b & In21+1k & 28.7 & 32.3 & 14.4 \\
         AVT-h & irCSN152 & IG65M & 25.5 & 28.1 & 12.8 \\
         \hline
         HORST & BNInc & In1k & 23.8 & 29.2 & 13.2 \\
         HORST-url & BNInc & In1k & 24.5 & 30.0 & 13.2 \\
    \end{tabular}
    \label{tab:ek100}
\end{table}

\subsection{Action Anticipation}
\subsubsection{Sampling.} We conduct action anticipation on EK55 and EK100. We process 14 frames from each clip, sampled with fixed stride of 0.25s (4 fps). We use $\tau_a$ to indicate anticipation time.

\subsubsection{Baselines.} We compare all baselines in RGB only modality. We include Deep Multimodal Regressor (DMR), \cite{DBLP:conf/cvpr/VondrickPT16}, TSN-based models MCE \cite{furnari2018leveraging} and ATSN \cite{atsn}, deep network trained with top-k classifier (SVM-Top3/5) \cite{berrada2018smooth}, Verb-Noun Marginal Cross Entropy (VNMCE) \cite{furnari2018leveraging}, and several LSTM variants, Encoder-Decoder LSTM (ED) \cite{DBLP:conf/bmvc/GaoYN17a}, Feedback Network LSTM (FN) \cite{DBLP:conf/wacv/GeestT18}, LSTM with Ranking Loss (RL) \cite{DBLP:conf/cvpr/MaSS16}, and LSTM with Exponential Anticipation Loss (EL) \cite{DBLP:conf/icra/JainSKSS16}. We also compare the state-of-the-art Rolling-Unrolling LSTM (RU) \cite{furnari2020rulstm}, ImagineRNN by predicting future feature \cite{wu2020learning}, and previous winners in EPIC-Kitchens anticipation challenge, ActionBanks in 2020 challenge \cite{sener2020temporal} and Anticipative Video Transformer (AVT) in 2021 \cite{girdhar2021anticipative}.

\subsubsection{Quantitative Results.}
Table~\ref{tab:ek55} reports top-5 accuracy at different anticipation times $\tau_a$, and presents top-5 accuracy and top-5 mean recall for each verb, noun and action at $\tau_a = 1\textit{s}$. Without any task specific design, HORST obtains 30.69\% top-5 action accuracy at $\tau_a = 1\textit{s}$. By equipping it with an unrolling classifier initialized from spatial-temporal states, $e_t$ and $h_t$ in equation~\eqref{eq:encoded_x}, top-5 accuracy boosts to 31.58\%, which is +0.75\% higher than state-of-the-art RULSTM. More comparisons on top-1 and top-5 accuracy are included in Table~\ref{tab:ek55_top1}. Under the same BN-Inception backbone, HORST achieves competitive top-5 accuracy over previous competition winners (ActionBanks, AVT), and comes with +0.9\% top-5 improvements when integrating the unrolling classifier. We observe lower top-1 scores, this may be due to cases where input observations are spanned over the previous action duration and could mislead the predictions by irrelevant frames with similar scenes or objects. Such overlapping causes distraction with negative impact on target action prediction, 
and the unrolling classifier may even amplify it. The same behavior is not found in early recognition.
 
Table~\ref{tab:ek100} shows quantitative results on EK100 dataset. AVT with irCSN152 backbone does not perform well on EK100, but is the strongest configuration in EK55. In contrast, AVT with AVT-b backbone has the best performance on EK100 but not on EK55. Such discrepancy can be the recall is used in evaluation, which reflects the par class performance but is hardly captured in top-1/5 accuracy measurements. HORST(-url) achieves the overall balanced performance on both EK55 and EK100. Although lower verb recall is observed, the noun and action are at the top level. We marked this difference as the direction to further develop upon our spatial-temporal attention design in future work.
 
\subsubsection{Visualization Analysis.}
Figure~\ref{fig:ek55_visa} shows an example with correct prediction on \texttt{Wash Board}. We can see the model only referencing the necessary states, which demonstrates that temporal attention is capable of skipping some specific frames. Spatial attention spots the hand movement successfully. Figure~\ref{fig:ek55_visb} reveals a challenging example labeled \texttt{Throw Onion}. The model struggles between \texttt{onion} and \texttt{skin} in top-5 noun predictions, by failing to locate the chopped onions on the cutting board shown in the very first two frames in the figure. Instead, the human hand and the onion skins are spotted, yielding the inaccurate, but plausible, top-5 predictions.

\section{Conclusion}
We have presented HORST, 
a Transformer-style architecture for video action prediction. Our model is simple, lightweight, and has a transparent design. Its spatial-temporal decomposition of self-attention  exposes visual explanations of model behaviors. HORST matches state-of-the-art performance on anticipation, and achieves superior performance on early recognition by a larger margin.

{\small
\bibliography{aaai22}
}


\clearpage
\setcounter{table}{7}
\setcounter{figure}{5}
\setcounter{equation}{11}
\setcounter{secnumdepth}{2}
\appendix

\section{Details on Classifier Designs}

In the following, we use \texttt{FC}$(v)$ to denote a linear (fully-connected) transformation on vector $v$. The parameters of different \texttt{FC}'s are distinguished by their subscripts.
 
\subsection{Early Action Recognition}

For early action recognition on SSv2 dataset, only the single action label for each sample is available. We took global average pooling followed by a flatten operator to obtain a 1D vector of $y_t$ (eq. (8) in the main paper), which is $\overline{y_t}$. The flattened 1D vector is then fed through a fully-connected layer, \texttt{FC}$(\overline{y_t})$, to obtain the action logits. The cross-entropy loss is deployed to compare between the action logit and the ground-truth action label at the last timestep, in both 25\% and 50\% settings.
 
\subsection{Anticipation}
\label{sec:app:cls}

As above, we took global average pooling followed by a flatten operator to obtain a 1D vector of $e_t$ as $\overline{e_t}$, and do the same on $\text{STATT}_t$ and $y_t$ as $\overline{\text{STATT}_t}$ and $\overline{y_t}$ (see eq. (7)(8) for the variables). The anticipation classifier is computed by:
 
\begin{eqnarray}
\chi_{a} &=& [\overline{e_t}; \overline{y_t}; \overline{\text{STATT}_t}] \label{eq:supp:ca}\\
\chi_{v} &=& \overline{\text{STATT}_t} \label{eq:supp:cv}\\
\chi_{n} &=& \overline{e_t} \label{eq:supp:cn}
\end{eqnarray}
and
 
\begin{eqnarray}
\text{logit}_{a} &=& \texttt{FC}_a(\chi_{a})\\
\text{logit}_{v} &=& \texttt{FC}_v(\chi_{v} + \texttt{FC}_{a2v}(\text{logit}_{a}))\label{eq:supp:logitv}\\
\text{logit}_{n} &=& \texttt{FC}_n(\chi_{n} + \texttt{FC}_{a2n}(\text{logit}_{a}))\label{eq:supp:logitn}
\end{eqnarray}
where the $\text{logit}_{a}$ is also served as biases to $\text{logit}_{v}$ and $\text{logit}_{n}$ for explicitly building the relationship of action to the verb and noun, and also to back-prop the supervision signals more fluently. 
 
The cross-entropy loss is used to compare between logits and ground-truth labels. The overall loss function is the summation over the individual loss of each action, verb, and noun, for every anticipation time.
 
\subsection{Anticipation with Unrolling}
\label{sec:app:cls-url}

HORST-url is shown in table 5-7 in the main paper with the unrolling classifier. The unrolling LSTM is performed on the final HORST outputs. The hidden and memory states of the unrolling LSTM are initialized by the $\overline{e_t}$ and $\overline{\text{STATT}_t}$,

\begin{eqnarray}
\xi_{v} &=& \texttt{ReLU}(\texttt{FC}_{\xi_v}(\overline{e_t} + \overline{\text{STATT}_t}))\\
\xi_{n} &=& \texttt{ReLU}(\texttt{FC}_{\xi_n}(\overline{e_t} + \overline{\text{STATT}_t}))\\
feat &=& \texttt{ReLU}(\texttt{FC}_{feat}([\xi_v; \xi_n]))\\
c^{url}_0 &=& \texttt{FC}_c(feat)\\
h^{url}_0 &=& \texttt{FC}_h(feat)
\end{eqnarray}
the states are unrolled T steps, where $T=\tau_a \times \text{fps}$, with the consistent input $\overline{y_t}$,

\begin{eqnarray}
(h^{url}_T, c^{url}_T) &=& \texttt{LSTMCell}(\overline{y_t}, (h^{url}_{T-1}, c^{url}_{T-1}))
\end{eqnarray}
We leverage the final unrolling output $h^{url}_T$ for further logits calculation, with the following inputs:

\begin{eqnarray}
\chi_{a} &=& h^{url}_T\\
\chi_{v} &=& \xi_v\\
\chi_{n} &=& \xi_n
\end{eqnarray}
Note that we supervise (verb, noun) on \eqref{eq:supp:logitv} and \eqref{eq:supp:logitn} with the inputs $\xi_v$ and $\xi_n$ to regularize the representation in the unrolling process.

\section{Training Time}
For early recognition on SSv2 dataset, the model of four HORST layers with hidden states set to [128, 256, 512, 512] is deployed. The total training time is about 25 hours per run on 25\% setting, this time is doubled in the 50\% setting. For action anticipation on EK55 dataset, we train two HORST layers with the hidden state set to 1024 on top of freezed BN-Inception backbone, the training needs to take about 50 hours per run. We set hidden state to 512 on EK100 to keep roughly the same computation budgets. All are conducted on 4 $\times$ NVIDIA V-100 GPUs.


\section{Code}
We present a PyTorch implementation of HORST layer in Figure \ref{fig:supp:impl}. We also include code files of this implementation (\texttt{horst.py}), and of the classifiers described in Sec.~\ref{sec:app:cls} (\texttt{cls-epic.py}) and Sec.~\ref{sec:app:cls-url} (\texttt{cls-url-epic.py}). These codes where used to produce the results reported in the main paper. All the codes are released at \href{https://github.com/CorcovadoMing/HORST}{\texttt{https://github.com/CorcovadoMing/HORST}}

\begin{figure*}
{\footnotesize
\begin{minted}{python}
import torch
from torch import nn
import torch.nn.functional as F


class SpatialFilter(nn.Module):
    def __init__(self):
        super().__init__()
        self.f = nn.Sequential(nn.Conv2d(2, 1, 3, 1, 1), nn.Sigmoid())
    
    def forward(self, x):
        # x: (b, c, h, w)
        # Implementation of Eq. (4) 
        avg_pool = x.mean(1, keepdim=True)
        max_pool = x.max(1, keepdim=True)[0]
        return self.f(torch.cat([avg_pool, max_pool], dim=1))

    
class STATT(nn.Module):
    def __init__(self, channels):
        super().__init__()
        self.sf_Q = SpatialFilter()
        self.sf_K = SpatialFilter()
    
    def forward(self, Q, K, V):
        # q: (1, b, c, h, w)
        # k: (s, b, c, h, w) 
        # v: (s, b, c, h, w)

        f_Q = self.sf_Q(Q.reshape(Q.size(0)*Q.size(1), *Q.shape[2:])) \
                  .reshape(*Q.shape[:2], 1, *Q.shape[3:])
        f_K = self.sf_K(K.reshape(K.size(0)*K.size(1), *K.shape[2:])) \
                  .reshape(*K.shape[:2], 1, *K.shape[3:])
        
        # Spatial branch - Implementation of Eq. (5)
        f_KQ = (f_K * Q).mean([-1, -2]) # \hat{Q} in Eq. (5)
        S = torch.sigmoid(torch.einsum('sbc,sbcd->sbd', f_KQ, K.reshape(*K.shape[:3], -1)))
        S = S.reshape(*V.shape[:2], 1, *V.shape[-2:]).expand_as(V)
        V = V * S # Eq. (7) - spatial part
        
        # Temporal branch - Implementation of Eq. (6)
        Q = (Q * f_Q).reshape(*Q.shape[:2], -1) 
        K = (K * f_K).reshape(*K.shape[:2], -1) 
        T = torch.einsum('lbd,sbd->lsb', Q, K)
        T = F.softmax(T * (Q.size(-1) ** -0.5), dim=1)
        V = torch.einsum('lsb,sbchw->bchw', T, V) # Eq. (7) - temporal part
        return V
\end{minted}
}
\caption{PyTorch implementation of our spatial-temporal decomposition of self-attention in HORST. It includes Eq.~(4-7) of the main paper.}
\label{fig:supp:impl}
\end{figure*}

\end{document}